\documentclass[letterpaper, 10 pt, conference]{ieeeconf}

\IEEEoverridecommandlockouts                              
\usepackage{amsmath,amsfonts}
\usepackage{algorithmic}
\usepackage{algorithm}
\usepackage{array}
\usepackage[caption=false,font=normalsize,labelfont=sf,textfont=sf]{subfig}
\usepackage{textcomp}
\usepackage{stfloats}
\usepackage{url}
\usepackage{verbatim}
\usepackage{graphicx}
\usepackage{cite}
\usepackage[T1]{fontenc}
\usepackage{booktabs}
\usepackage{multirow}
\usepackage{lscape}
\usepackage[colorlinks,
            linkcolor=blue,
            anchorcolor=blue,
            citecolor=blue,
            urlcolor=blue]{hyperref}
\usepackage{yhmath}
\usepackage{diagbox}
\usepackage{tcolorbox}

\title{\LARGE \bf
\textit{Follow Everything:} Goal-Aware Adaptation and Graph-Based Planning Towards Arbitrary Leader Following
}

\author{Qianyi Zhang$^{1}$, Shijian Ma$^{2}$, Boyi Liu$^{3}$, Jingtai Liu$^{1}$, Jianhao Jiao$^{^{\dagger}, 4,5}$, and Dimitrios Kanoulas$^{\dagger},^4$
\thanks{$^1$Institute of Robotics and Automatic Information System, Nankai University, China. 
$^2$Centre for Data Science, University of Macau, China. $^3$Electrical and Computer Engineering Department, Hong Kong University of Science and Technology, China. $^4$Department of Computer Science, University College London, UK. $^5$Department of Aeronautical and Aviation Engineering, The Hong Kong Polytechnic University, Hong Kong, China.}
\thanks{[$^{\dagger}$] The corresponding authors. This work was done during Qianyi, Shijian, and Boyi's visit at University College London. This work is supported by UK Research and Innovation Future Leaders Fellowship (RoboHike, Grant No. MR/V025333/1) and in part by the National Natural Science Foundation of China under Grant 62573244.}
}

\begin{document}

\maketitle
\thispagestyle{empty}
\pagestyle{empty}


\begin{abstract}
Enabling robots to robustly follow leaders supports tasks such as carrying supplies or guiding customers. While existing methods often fail to generalize to arbitrary leaders, and struggle when the leader temporarily leaves the robot’s field of view, this work presents a unified framework to address both challenges. First, a segmentation model replaces traditional category-specific detection models, allowing the leader to be of any shape or type. To improve robustness, a distance frame buffer is designed to store high-confidence leader embeddings across distance intervals, accounting for the unique characteristics of leader-following tasks. Second, a goal-aware adaptation mechanism is designed to govern robot planning states based on the leader's visibility and motion, complemented by a graph-based planner that generates candidate trajectories for each state, ensuring efficient following with obstacle avoidance.
Simulations and real-world experiments with a legged robot follower and diverse leaders in indoor and outdoor settings demonstrate an improved follow success rate of 75.1\%, a reduced visual loss of 13.1\%, a fewer collisions of 65.1\%, and a shorter average distance of 1.3 m.
Visit \href{https://follow-everything.github.io}{https://follow-everything.github.io} for the video and code.
\end{abstract}

\section{Introduction}
\begin{figure}[thb]
	\centering
	\includegraphics[width=3.4in]{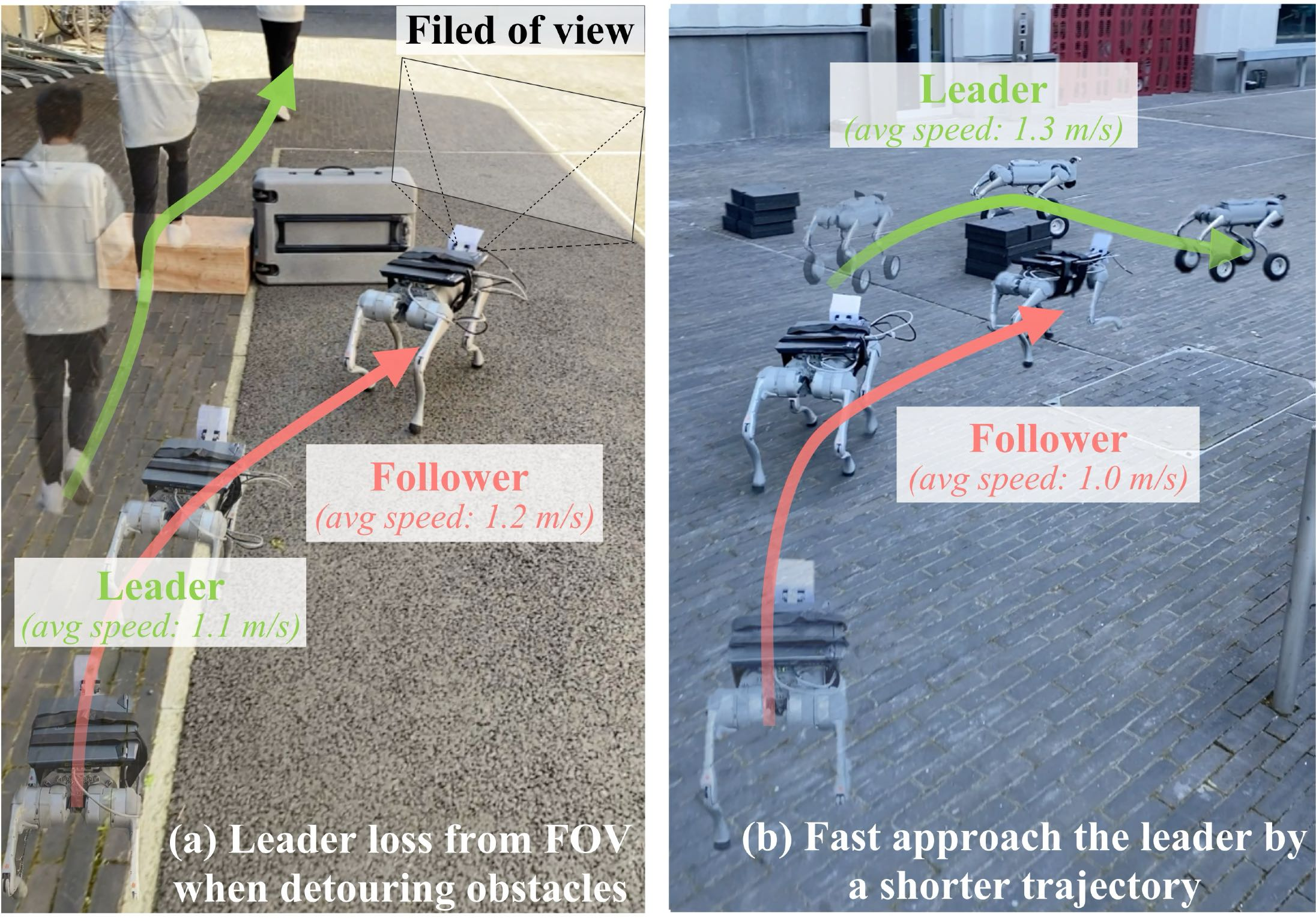}
	\caption{Illustration of the challenges in leader-following tasks. (a) When the leader steps over a box that the follower must detour around, the leader leaves the follower’s field of view. (b) The follower is expected to follow the leader via a shorter trajectory, not by replicating the leader’s past path.}
	\label{pic_motivation}
\end{figure}

Recent advances in visual perception, robot navigation, and large models have enabled robots to follow leaders with increased safety and intelligence~\cite{10909198}. 
Serving as powerful assistants, robots can follow explorers to carry more supplies, guide customers and introduce products in shopping centers, help police patrolling, etc.~\cite{rpl1,rpl2,rpl3,rpl4}.
Despite extensive efforts to improve the stability of leader following~\cite{wang2025trackvla, 2023_yolo, 2021_lidar, 2023_multimodal} and the smoothness of robot motion~\cite{2024_follow_anything, 2025_MPC_TRO, gaze, zhang2021efficient, LiuDiPPeR2024, han2025neupan}, existing approaches still either fail to re-identify the leader after it temporarily leaves the robot’s field of view (FOV) and later reappears (see Fig.\ref{pic_motivation}a), or they cannot approach the leader in the most efficient manner (see Fig.\ref{pic_motivation}b).
This paper breaks down these challenges from both the perception and planning perspectives and proposes the distance frame buffer, goal-aware adaptation, and graph-based planner to address them.

The challenge on the perception side involves re-identifying the leader after it exits and re-enters the follower’s FOV. One specific approach for leader-following tasks is to have the leader carry a beacon, such as infrared antennas~\cite{2022_laser} or wave emitters~\cite{2023_wave,2024_wave}, but requiring the leader to carry additional equipment is often impractical and not generalizable. Therefore, another widely used solution is to leverage detection models based on cameras~\cite{2023_yolo}, LiDAR~\cite{2021_lidar}, multimodal sensors~\cite{2023_multimodal}, sonar~\cite{2021_ultrasonic}, and so on. However, since their training labels correspond to object categories rather than object features, when the leader (typically a pedestrian) remains in the follower’s FOV, its ID remains stable due to the help of the Kalman Filter~\cite{2015_kalman_filter}. However, when the leader leaves and reappears, a new ID will be assigned to it. This leads to the expected leader having the original ID permanently lost.

\begin{figure*}[!t]
	\centering
	\includegraphics[width=7.0in]{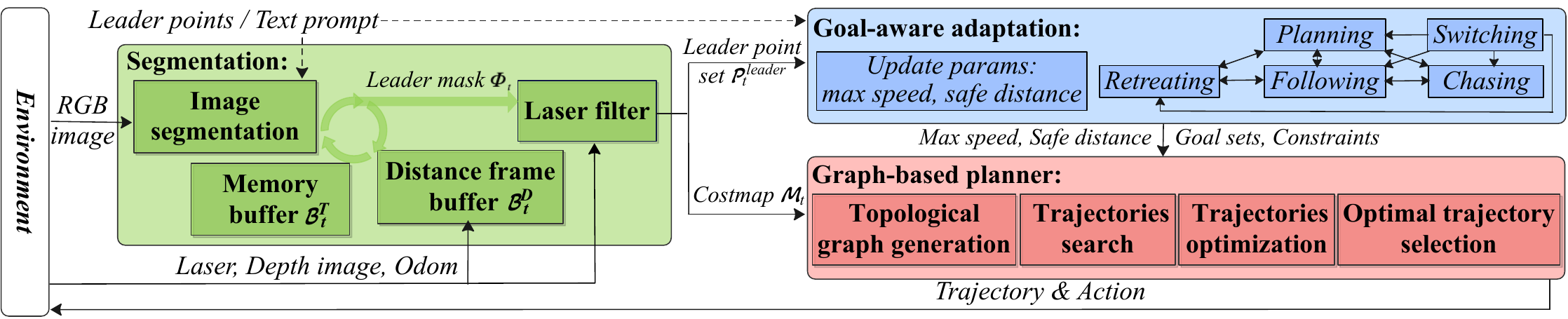}
	\caption{Illustration of the proposed \textit{follow everything} framework. Given a leader prompt and an RGB image, the leader mask is segmented with the aid of both a memory buffer and a distance frame buffer. The segmented leader is then filtered from raw laser point clouds to produce a leader point set, which is passed to the goal-aware adaptation to dynamically adjust parameters and provide goal sets and constraints according to the follower-leader interaction. Given the costmap, goal sets, and constraints, the graph-based planner searches, optimizes, and selects an optimal trajectory for the follower to execute.}
	\label{pic_framework}
\end{figure*}

An alternative solution is segmentation models~\cite{2024_evf_sam2} that use features as labels: with an initial feature input, the model tracks and bootstrap-updates the leader’s features continuously, enabling re-identification upon its reappearance~\cite{2024_follow_anything}. 
These models are not category-limited and can track arbitrary-shaped leaders (humans, vehicles, robots, etc.). However, a limitation of existing segmentation datasets is that they often involve a fixed camera position (follower) with the leader's movement, location and size constantly changing in the video, where bootstrapping is beneficial.

However, in the leader-follower task, the follower is also in motion, making it easy for the leader to leave the follower’s FOV, as illustrated in Fig.~\ref{pic_motivation}(a). Just before leaving, the leader often appears on the edge of the robot’s FOV, resulting in partial and low-quality features. Despite their poor quality, these features can still dominate the bootstrapping process, since the older features that capture the full-body shape of the leader differ significantly from the current appearance of the leader~\cite{2024_follow_anything}. This can mislead the robot to bootstrap in the wrong direction. As a result, when the leader reappears later, the robot may not be able to retrace the leader accurately, even if the full leader is in front of it.
To address this issue, this work introduces the distance frame buffer to SAM2 (see Sec.~\ref{m1}), which divides the distance ranges between the follower and leader into equal intervals and maintains the historical leader feature with the highest confidence in each distance interval, greatly improving the ability to retrace the leader after it reappears.

The challenge on the planning side involves obstacle avoidance and robust motion planning when the relationship between the leader and the follower is dynamic. While most existing approaches simply treat the leader’s position as a target and use PID~\cite{2024_follow_anything}, MPC~\cite{2025_MPC_TRO}, sampled trajectories~\cite{gaze}, optimized trajectories~\cite{zhang2021efficient}, or learning-based methods~\cite{han2025neupan} to plan trajectories for the follower (robot), they often lack robustness and fail to consider the various interactive states that arise from the leader-follower relationship.
For example, (a) when the leader is within the follower’s FOV and close by, the main goal of the follower is to avoid obstacles while following, a scenario that frequently occurs when a legged robot follows a UAV leader, as shown in Fig.~\ref{pic_sim2}, where the UAV can fly over short obstacles that the ground robot must detour around. (b) When the leader is within the follower’s FOV but at a greater distance, the main goal of the follower is to quickly approach the leader, which requires the robot to maintain high speed and choose optimal time trajectories to avoid increasing the risk of losing the leader due to excessive distance, as shown in Fig.~\ref{pic_sim4}. (c) When the leader retreats, the follower should also retreat accordingly while maintaining a safe margin to avoid collisions with the leader or losing sight of it due to too-close proximity, as shown in Fig.~\ref{pic_sim3}d. (d) When the leader leaves the follower’s FOV, the follower should intentionally approach the last known position and orientation of the leader to recover it as quickly as possible, as illustrated in Fig.~\ref{pic_exp}f-g. (e) A robust model should also be able to replace the leader if necessary, improving the practicality. 
To realize these robust following capabilities, this work designs a goal-aware adaptation (see Sec.\ref{m2}) under the proposed \textit{follow everything} framework, which dynamically adjusts the robot’s speed and the desired safe distance between the leader and the follower, and adaptively determines the follower’s state to provide unique goals and 
constraints for the subsequent trajectory planner.

As for the planner, the primary concern is to find a time-optimal trajectory for the follower while adapting to the goals and constraints provided by the goal-aware adaptation. Building on previous work~\cite{2024_STCTEB, 2025_GATEB}, the environment can be modeled as a graph, with obstacles clustered as nodes and their shortest connections as edges. Given that each node has different detour directions, a series of candidate trajectories with different meanings~\cite{2025_tro_homo, 2010_homo} can be generated. Besides kinematic constraints, the goal sets and unique constraints provided by goal-aware adaptation are incorporated into the optimization. The trajectory with the shortest time to reach the goal is selected as the best trajectory to execute.
This graph-based planner (see Sec.~\ref{m3}) enables the follower to approach the leader efficiently, rather than simply replicating their historical paths, as shown in Fig.~\ref{pic_motivation}(b).

In summary, this work presents a unified framework for robust leader-following and obstacle avoidance. 
The main contributions are as follows:
\begin{itemize}
    \item A segmentation module with a distance frame buffer is introduced to enable everything to be a leader.
    \item A goal-aware adaptation strategy is proposed to enable state switching for robust leader-following.
    \item A graph-based trajectory planner is developed to produce candidate trajectories for safe following.
\end{itemize}

\begin{figure*}[!t]
	\centering
	\includegraphics[width=7.0in]{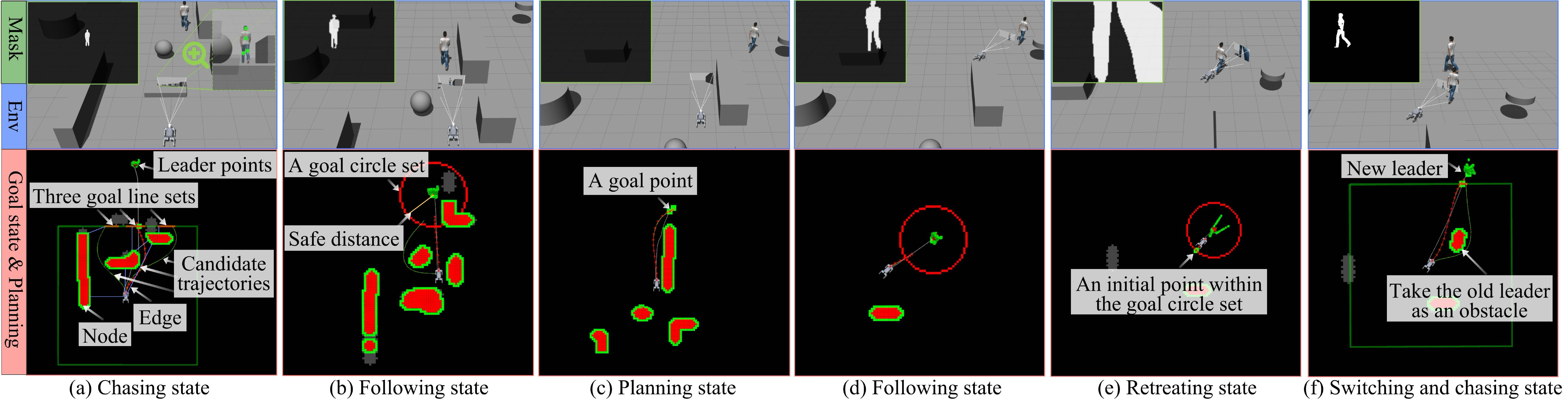}
	\caption{Illustration of goal-aware adaptation and graph-based planner in a demo. (a) The leader is segmented after selecting points. \textit{Chasing} is activated as the leader is far away. (b) \textit{Following} is triggered when the robot gets closer, maintaining a safe distance. (c) \textit{Planning} is activated when the leader exits the robot's FOV, guiding the robot to the last known leader pose. (d) \textit{Following} resumes once the leader is visible again. (e) \textit{Retreating} is triggered when the leader steps back. (f) A new prompt "follow the person on the left side" activates \textit{Switching}, followed by \textit{Chasing} as the new leader is distant.}
	\label{pic_method}
\end{figure*}

\section{Methodology}
\subsection{Problem Formulation}
Given the RGB image \( I_t \) at timestamp \( t \), the leader is segmented into an embedding \( \eta_t \) and a mask \( \phi_t \) with the assistance of a temporal memory buffer \( \mathcal{B}_t^T \) and a distance frame buffer \( \mathcal{B}_t^D \). Using the depth image, the leader mask \( \phi_t \) is transformed into a point set \( \mathcal{P}^{leader}_t = \{\textbf{p}_1, \textbf{p}_2, ..., \textbf{p}_m\} \), where each point \( \textbf{p}_i = (x, y, z) \) represents the 3D coordinates of the leader.
The leader's position \( \bar{\textbf{p}}^l_t \) is obtained by averaging the point set \( \mathcal{P}^{leader}_t \), and the leader's velocity \( \bar{\textbf{v}}^l_t \) is estimated based on the positional difference over a fixed time interval.
Given the robot's position \( \textbf{p}^r_t \) and the leader's states \( \bar{\textbf{p}}^l_t \) and \( \bar{\textbf{v}}^l_t \), the goal-aware adaptation mechanism determines the robot’s maximum speed \( V_t^{max} \) and the safe distance \( D_t \) between the robot and the leader. Additionally, the robot's goal is provided as a series of sets \( \{\mathcal{G}_t^0, \mathcal{G}_t^1, ..., \mathcal{G}_t^n\} \), each associated with the desired orientations.
A topological graph is constructed based on a costmap \( \mathcal{M}_t \), where obstacle groups are defined as nodes \( \mathcal{N}_i \) and the shortest connections between groups are defined as edges \( \mathcal{E}_{i,j} \). This graph generates several generalized trajectories \( \mathcal{T}_i \), which are then transformed into normal trajectories \( \tau_i \).
By optimizing these trajectories with constraints imposed by the robot's kinematics, goal sets $\mathcal{G}_t^i$, maximum speed $V_t^{max}$, and safe distance $D_t$, the optimal trajectory is selected and executed.

\subsection{Leader Segmentation with Distance Frame Buffer} \label{m1}
Given an initial RGB image $I_0$ and a set of leader point prompts (often manually selected; see Fig.~\ref{pic_method}a), the segmentation model~\cite{2024_evf_sam2} expands the prompts to the entire leader, producing a leader embedding $\eta_0$ and a leader mask $\phi_0$.
Then, at each subsequent timestep $t$, the top $n$ historical leader embeddings with the highest confidence scores are maintained and updated in a temporal memory buffer $\mathcal{B}_t^T$: 
\begin{equation}
\mathcal{B}^{T}_t = \{\eta^{1}, ...\}_{n_1},  \;\eta^i = \mathop{\arg\max}\limits_{\forall \eta_t \; \cap \; \eta_t \notin [\eta^1, ..., \eta^{i-1}] } \; S(\eta_t),
\end{equation}
where $\forall \eta_t$ contains all historical embeddings and $S(\eta_t)$ is the confidence score.
The distance frame buffer \( \mathcal{B}_t^D \) considers the spatial dimension.
It has a length of \( n_2 \), with each element \( \eta^i \) storing the leader embedding with the highest confidence score from historical leader embeddings within the distance range \( [(i-1)\Delta d, i \Delta d] \) between the follower and the leader: 
\begin{equation}
    \mathcal{B}_t^D=\{\eta^1,... \}_{n_2},\eta^i = \mathop{\arg\max}\limits_{\forall \eta_t \; \cap \; d(\eta_t) \in [(i-1)\Delta d, i \Delta d] } \; S(\eta_t),
\end{equation}
where \( \Delta d \) is a distance interval, and the distance between the follower and the leader is defined as \( d(\eta_t) = |\bar{\mathbf{p}}_t^l - \mathbf{p}_t^f| \).
Given the current image $I_t$ and the two buffers, the current leader embedding $\eta_{t}$ and mask $\phi_t$ can be generated: 
\begin{equation}
	\eta_{t}, \phi_{t} = \text{Segmentation}(I_{t}, \mathcal{B}^T_t, \mathcal{B}^D_t).
\end{equation}

The introduction of the distance frame buffer enables the segmentation model to isolate embeddings based on distance, reducing the contamination of the entire buffer when bootstrap occurs in the wrong direction, such as when the leader is partially observable or in the process of leaving the follower's FOV. Its effectiveness can be seen in Table.~\ref{table}, where our Follow Everything achieves the highest follow success rate metric compared to Alaa and FE-N-DFB.

\subsection{Graph-based Planner} \label{m3}
The graph-based planner is based on time-optimal optimization~\cite{2017_RAS_TEB} and provides a unified framework to incorporate various constraints of goal-aware adaptation, making the decision-making of the robot consistent even when motion states frequently alternate. 
Filtering the leader points $\mathcal{P}_{leader}$ from the laser scan, the remaining points form a binary costmap $\mathcal{M}_t$. Following prior work~\cite{2024_STCTEB, 2025_GATEB}, adjacent obstacle points in $\mathcal{M}_t$ are clustered into obstacle groups, each outlined by a boundary. 
A topological graph is established with the obstacle groups as nodes $\mathcal{N}_i$. For any two boundaries with indices $i$ and $j$, their shortest collision-free connection is defined as an edge $\mathcal{E}_{i,j}$ (see Fig.~\ref{pic_method}a) of the graph.
Given the goal point or goal sets $\{\mathcal{G}_t^0, \mathcal{G}_t^1, ...\}$ from the goal-aware adaptation, which 
will be detailed later, several generalized trajectories can be generated over the topological graph using depth-first search, with each one $\mathcal{T}$ following the form:
\begin{equation}
    \mathcal{T} = \textbf{p}_t^r \to \mathcal{E}_{r,i} \to \wideparen{\mathcal{N}}_i \to \mathcal{E}_{i,j} \to \wideparen{\mathcal{N}}_j \to \ldots \to \mathcal{G}_k.
\end{equation}

Since each obstacle node $\mathcal{N}_i$ can be detoured in clockwise or counterclockwise directions, a generalized trajectory $\mathcal{T}$ containing $n$ nodes can produce up to $2^n$ distinct trajectories $\tau$. 
To eliminate redundant trajectories that represent the same detour behavior, the trajectory homotopy classes~\cite{2025_tro_homo, 2010_homo} are applied. 
The remaining trajectories are then optimized in parallel. For each trajectory $\tau = \{ \textbf{p}_0, \textbf{p}_1, ...,\textbf{p}_m\}$, 
it follows the time-optimal optimization framework: 
\begin{align}
    \tau &= \mathop{\arg\min}\limits_{\{\textbf{p}_1,...,\textbf{p}_m \}} \sum_{t=1}^{m} ||\textbf{p}_i - \textbf{p}_{i-1}||_2 / V_t^{max}.  \label{eq_time} \\
    & \text{s.t.} \quad \mathcal{O}_{dynamic}(\textbf{p}_i) = 0, \quad
    \mathcal{O}_{static}(\textbf{p}_i) = 0, \\
    & \quad \quad \mathcal{H}(\textbf{p}_i, \textbf{p}_{i+1}, \textbf{p}_{i+2}) = 0, \quad  \mathcal{A}(\textbf{p}_i, \textbf{p}_{i+1}, \textbf{p}_{i+2}) \leq 0, \\
    & \quad \quad \mathcal{I}(\textbf{p}_m, \mathcal{G}_{i}) \leq 0, \quad \mathcal{V}(\textbf{p}_i, \textbf{p}_{i+1}, V_t^{max}) \leq 0, \label{equ_goal_aware}
\end{align}
where $\mathcal{O}_{dynamic}(\cdot)$ and $\mathcal{O}_{static}(\cdot)$ represent the constraints for avoiding dynamic~\cite{2024_STCTEB} and static~\cite{2025_GATEB} obstacles, respectively. 
$\mathcal{H}(\cdot)$ and $\mathcal{A}(\cdot)$ denote the kinematic and acceleration constraints~\cite{2017_RAS_TEB}, respectively. 
$\mathcal{I}(\cdot)$ and $\mathcal{V}(\cdot)$ are the goal and velocity constraints, both of which are provided by the goal-aware adaptation and will be detailed in the next subsection.

All the trajectories are evaluated, and the one $\tau^*_t$ with the lowest cost is selected and executed by the robot: 
\begin{equation}
	\begin{gathered}
		\tau^*_t = \mathop{\arg\min}\limits_{\tau} f(\tau^*_{t-1}, \tau) g(\tau), \\
		f(\tau^*_{t-1}, \tau) = (1-\alpha) + \alpha \sum_{i=1}^l \frac{val_i(\tau^*_{t-1}) - val_i(\tau)}{l},
	\end{gathered}
\end{equation}
where $f(\tau^*_{t-1}, \tau)$ evaluates the similarity between the previous optimal trajectory $\tau^*_{t-1}$ and the current candidate $\tau$, and $g(\tau)$ evaluates the time cost of the trajectory, as defined in Equ.(\ref{eq_time}).
In a scenario with $l$ obstacle groups, a trajectory is assigned a homology signature $\{val_1, val_2, ..., val_l\}$, where each $val_i \in \{1, -1\}$ indicates whether the trajectory detours the $i$-th obstacle group in a clockwise or counter-clockwise direction, respectively.
$\alpha$ is a scaling factor.
This design enhances the robustness of trajectory selection, even when the goal sets dynamically change due to goal-aware adaptation.


\subsection{Goal-aware Adaptation} \label{m2}
The goal-aware adaptation governs the goal constraint $\mathcal{I}(\cdot)$ and velocity constraint $\mathcal{V}(\cdot)$ during trajectory optimization, balancing safety in avoiding obstacles and efficiency in following the leader.
The transition between goal states is determined by conditions, including whether the leader is within the robot's FOV, whether it lies within the costmap, the robot-leader distance, and whether a new leader is assigned. See Fig.~\ref{pic_framework} for possible transitions between states. 

\textbf{Chasing State} (Fig.~\ref{pic_method}a) is activated when the leader is within the robot’s FOV but outside the costmap range, where the robot’s primary task is to navigate around nearby obstacles and chase the leader as quickly as possible. 
Unlike conventional methods that select the local goal as the intersection point between the global path and the costmap edge, the chasing state extends this point along the costmap edge to form a goal line of length $L_t$, which is proportional to the distance between the robot $\textbf{p}_{r}$ and the leader $\bar{\textbf{p}}_{l}$:
\begin{equation}
    L_t = \alpha (|\textbf{p}_t^r - \bar{\textbf{p}}_t^{l}|-\frac{W_{map}}{2}),
\end{equation}
where $W_{map}$ is the costmap width and $\alpha$ is a scaling factor. 
\begin{figure*}[!t]
	\centering
	\includegraphics[width=7.0in]{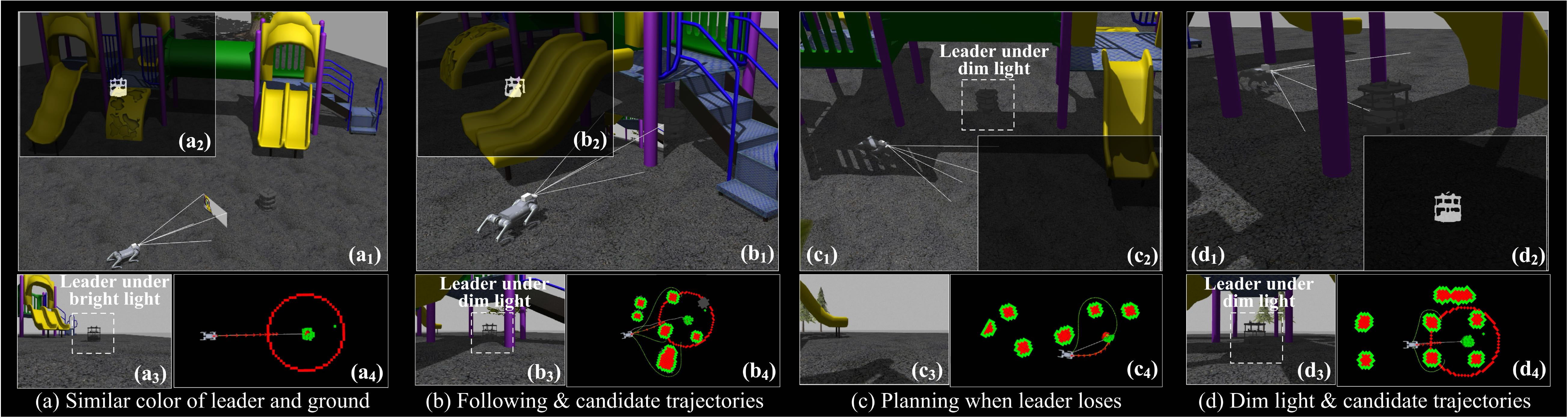}
	\caption{Simulation for following a mobile robot in a playground. (a) The leader can be robustly segmented even when its color closely resembles the ground. (b) Multiple candidate trajectories are generated for the robot to choose from. (c) When the leader is lost, the planning state enables the follower to actively search for the leader instead of awkwardly stopping in place. (d) Even under dim lighting conditions, the leader can still be robustly segmented.}
	\label{pic_sim1}
\end{figure*}

The whole goal line is separated at the locations where it intersects with obstacles, resulting in several sub-goal line sets $\mathcal{G}_i=\{\textbf{p}^g_1, \textbf{p}^g_2, ..., \textbf{p}^g_{n_i} \}$, which are then added as additional nodes to the topological graph along with their corresponding connections.
Therefore, the goal constraint $\mathcal{I}(\textbf{p}_m^{\tau}, \mathcal{G}_i)$ of Equ.(\ref{equ_goal_aware}) in this state is defined as: 
\begin{equation}
	\begin{gathered}
	\overrightarrow{\textbf{p}^g_1 \textbf{p}^g_{n_i}} \times  \overrightarrow{\textbf{p}^g_1, \textbf{p}^{\tau}_m} = 0, \\
	0 \leq \overrightarrow{\textbf{p}^g_1 \textbf{p}^g_{n_i}} \cdot  \overrightarrow{\textbf{p}^g_1, \textbf{p}^{\tau}_m} \leq |\overrightarrow{\textbf{p}^g_1 \textbf{p}^g_{n_i}}|^2, 
	\end{gathered}    
\end{equation}
where $\textbf{p}^g_1$ and $\textbf{p}^g_{n_i}$ denote the endpoints of a goal set $\mathcal{G}_i$, and the $\textbf{p}^{\tau}_m$ is the last point on the trajectory $\tau$. 
The goal constraint in this state allow the point to $\textbf{p}^{\tau}_m$ slides along the goal set during optimization so that a faster trajectory can be optimized and obtained. 

The velocity constraint $\mathcal{V}(\textbf{p}_i^{\tau}, \textbf{p}_{i+1}^{\tau}, V^{max})$ of Equ.(\ref{equ_goal_aware}) considers only the robot’s physical speed limit $V^{max}$ in this state. For each adjacent points, the constraint is defined as: 
\begin{equation}
    ||\textbf{p}_i^{\tau} - \textbf{p}_{i+1}^{\tau}||_2 / \Delta T \leq V^{max},
\end{equation}
where $\Delta T$ is a fixed time interval. In this case, the robot can approach the goal as quickly as possible.

\textbf{Following State} (Fig.~\ref{pic_method}b and Fig.~\ref{pic_method}d) is activated when the leader is within the robot's FOV and also within the current costmap. The primary task is to follow the leader while avoiding unnecessary detours, as in the experiment in Fig.~\ref{pic_exp}b.  
A Kalman filter is used to estimate the leader’s state, where the Normalized Innovation Squared (NIS)~\cite{2015_kalman_filter} value serves as an indicator of prediction uncertainty. Based on this, a safe distance between the follower and the leader is dynamically adjusted as:
\begin{equation}
D_t = \text{Clip}\left(\alpha \text{NIS}, \, D^{min}, \, D^{max} \right),
\end{equation}
where $D^{min}$ and $D^{max}$ are predefined lower and upper bounds, and $\alpha$ is a scaling factor. 
Instead of using goal line sets as in the \textit{chasing} state, candidate goal points are sampled along a circle centered on the leader, with a radius equal to the safe distance \( D_t \) in this state. These points are further segmented by obstacles into several sets of subgoal circles \( \mathcal{G}_i \). In this state, the goal constraint \( \mathcal{I}(\textbf{p}_m^{\tau}, \theta_m^{\tau}, \mathcal{G}_i) \) at Equ.(\ref{equ_goal_aware}) allows the last trajectory point \( \textbf{p}_m^{\tau} \) to slide along the goal circle set $\mathcal{G}_i=\{\textbf{p}^g_1, \textbf{p}^g_2, ..., \textbf{p}^g_{n_i} \}$, while constraining the orientation \( \theta_m^{\tau} \) to face the leader, thereby reducing the risk of losing visual contact:
\begin{equation}
\begin{gathered}
    \overrightarrow{\bar{\textbf{p}}_{l} \; \textbf{p}_1^g} \times \overrightarrow{\bar{\textbf{p}}_{l} \; \textbf{p}_m^{\tau}} \geq 0, \hspace{20pt} 
    \overrightarrow{\bar{\textbf{p}}_{l} \; \textbf{p}_m^{\tau}} \times \overrightarrow{\bar{\textbf{p}}_{l} \; \textbf{p}^g_{n_i}}\geq 0,    
    \\ 
    |\overrightarrow{\bar{\textbf{p}}_{l} \; \textbf{p}_m^{\tau}}|=D_t, \hspace{20pt}
    | \theta_m^{\tau}- atan(\bar{\textbf{p}}_{l} - \textbf{p}_m^{\tau}) | \leq \epsilon,
\end{gathered}
\label{following}
\end{equation}
where $\epsilon$ is a small threshold. 

The maximum speed of the robot $V_t^{max}$ in this state is determined by the relative distance between the robot $\textbf{p}_t^r$ and the leader $\bar{\textbf{p}}_t^l$, as well as the estimated velocity of the leader $\bar{\textbf{v}}_t^l$. 
The velocity constraint $\mathcal{V}(\textbf{p}_i^{\tau}, \textbf{p}_{i+1}^{\tau}, V^{max})$ of Equ.(\ref{equ_goal_aware}) reduces the robot’s physical speed limit $V^{max}$ to $V^{max}_t$, avoiding frequent stop-and-go behavior of the robot. For each adjacent points, the constraint is defined as: 
\begin{equation}
\begin{gathered}
V_t^{max} = \text{Clip}\left(\alpha_1 |\bar{\textbf{v}}_t^l| + \alpha_2 |\bar{\textbf{p}}_t^l - \textbf{p}_t^r|, \, 0, \, V^{max} \right), \\
    ||\textbf{p}_i^{\tau} - \textbf{p}_{i+1}^{\tau}||_2 / \Delta T \leq V^{max}_t,
\end{gathered}
\end{equation}
where $\alpha_1$ and $\alpha_2$ are scaling factors. The function $\text{Clip}(\cdot)$ limits the value within the physical kinematics of the robot.

\textbf{Planning state} (Fig.~\ref{pic_method}c) is activated when the leader is no longer within the robot’s field of view. This typically occurs when the robot is detouring around an obstacle or alternates with the \textit{following} state when the leader is partially occluded. In this state, the goal constraint $\mathcal{I}(\textbf{p}_m^{\tau},\theta_m^{\tau},\bar{\textbf{p}}_t^l)$ of Equ.(\ref{equ_goal_aware}) sets the trajectory endpoint $\textbf{p}_m^{\tau}$ and $\theta_m^{\tau}$ to be the last known position $\bar{\textbf{p}}_t^{l}$ and orientation $\bar{\theta}_t^{l}$ of the leader:
\begin{equation}
\begin{gathered}
    | \textbf{p}^\tau_{m} - \bar{\textbf{p}}_t^{l} | <=\epsilon, \hspace{20pt} | \theta_m^{\tau} - \bar{\theta}_t^{l} | <=\epsilon.
\end{gathered}    
\end{equation}

The velocity constraint $\mathcal{V}(\textbf{p}_i^{\tau}, \textbf{p}_{i+1}^{\tau}, V^{max})$ of Equ.(\ref{equ_goal_aware}) obeys the same way as that of \textit{chasing} state, so that the robot can quickly approach the leader to reduce the risk of losing the visual contact. 

\textbf{Retreating State} (Fig.~\ref{pic_method}e) is activated when the leader actively approaches the robot. The robot is expected to move backward to avoid potential collisions with the leader. The velocity constraint $\mathcal{V}(\textbf{p}_i^{\tau}, \textbf{p}_{i+1}^{\tau}, V^{max}_t)$ and goal constraints \( \mathcal{I}(\textbf{p}_m^{\tau}, \theta_m^{\tau}, \mathcal{G}_i) \) in this state are identical to those used in the \textit{following} state. The one closest to the robot is selected as the initial goal, and the endpoint of the trajectory is allowed to slide along the corresponding goal set $\mathcal{G}_i$. 

\textbf{Switching state} (Fig.~\ref{pic_method}f) is activated when a new leader is assigned, most commonly triggered by a large language model~\cite{2024_evf_sam2}. In this state, both the memory buffer and the distance frame buffer are reset based on the current input. The system then directly transitions to the appropriate subsequent state according to the relative relationship between the robot and the newly assigned leader.

\section{Simulation}
The simulation evaluates our Follow Everything in a Gazebo with a legged robot follower ($V^{max} = 1.5,\mathrm{m/s}$), using a Realsense D435i for segmentation and a Mid360 for point clouds.
Four scenarios are tested: a mobile robot as the leader in a playground, a UAV as the leader in a forest, a pedestrian as the leader in a factory, and a stop sign as the leader in a dynamic scenario.
In each test scenario, $10$ leader trajectories are recorded using ROS bag and replayed to ensure a fair comparison. Each trajectory is tested four times to minimize the effects of environmental noise, resulting in $40$ tests per scenario and $160$ tests in total. 
Four metrics~\cite{zhang2021efficient} are considered: (1) the \textit{follow success rate} (the percentage of runs where the robot maintains a safe distance and keeps the leader in view after the leader reaches its target), (2) the \textit{average leader loss time ratio} (the ratio of time the leader is out of view), (3) the \textit{collision rate} (the ratio of collisions over 160 tests), and (4) the \textit{average distance} (the average robot–leader distance, where larger values risk visual loss).
Our method, Follow Everything, is compared against three baselines: (1) Alaa~\cite{2024_follow_anything}, which segments the leader using SAM2 and plans trajectories with a PID controller; (2) FE-N-DFB, an ablation of our framework that does not incorporate the distance frame buffer; and (3) FE-N-GP, another ablation of our framework that replaces the graph-based planner with the MPC controller.
Further details regarding setups and parameters are available at the \href{https://follow-everything.github.io}{website}. 

\begin{figure*}[!t]
	\centering
	\includegraphics[width=7.0in]{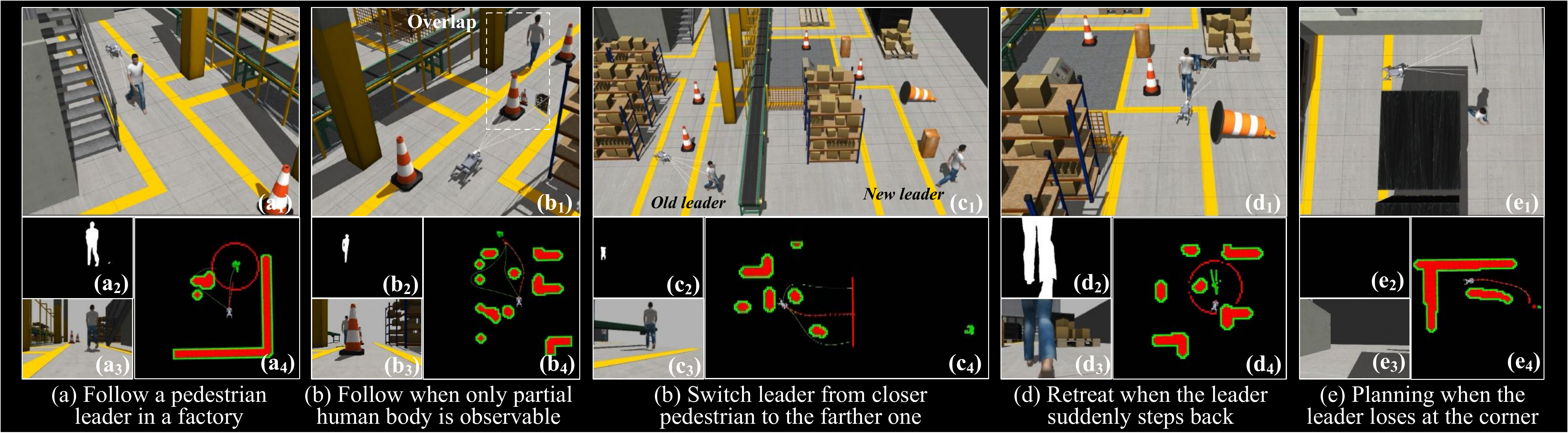}
	\caption{Simulation for following a pedestrian in a factory. (a–b) The robot follows the leader when partially observed. (c) Upon receiving a prompt specifying a new leader, the chasing state is activated, and the robot passes underneath the workstation to approach the new leader. (d) When the leader steps backward, the robot retreats to maintain a safe distance. (e) When the leader turns a corner and leaves the robot's FOV, the planning state is triggered.}
	\label{pic_sim3}
\end{figure*}

\textbf{Mobile robot leader in playground}. 
In this scenario, the leader is a mobile robot whose color closely resembles that of the ground, as shown in Fig.~\ref{pic_sim1}(a). A building is placed at the center of the scenario, where narrow gaps are just wide enough for both the leader and the follower to pass through.

This scenario presents three major challenges. The first challenge is whether the follower can robustly segment and track the leader under varying lighting conditions. As shown in Fig.~\ref{pic_sim1}(a), when the leader is outside the slide, the lighting is relatively bright. In contrast, inside the slide (Fig.~\ref{pic_sim1}b–d), the lighting becomes dim, which can significantly alter the visual features of the leader. This discrepancy often leads to detection or segmentation failures, even when the leader is directly in front of the follower. 
The SAM-based memory buffer continuously updates the leader embedding over time to adapt to illumination variations and mitigate the risk of tracking failure. By incorporating a distance frame buffer into Follow Everything, we further expand the original memory buffer to enable the model to leverage richer features for tracking, achieving the minimal average leader tracking loss time ratio of 10.7\%.

The second challenge is whether the robot can safely follow the leader. As shown in Fig.~\ref{pic_sim1}(b) and (d), the slide and pillars create several impassable regions. The robot is therefore unable to move directly toward the leader and must instead follow while simultaneously avoiding obstacles. Our graph-based planner addresses this by generating multiple candidate trajectories with distinctive meanings, from which the safest and fastest path can be selected.

The third challenge involves re-identifying the leader after it disappears from the robot’s FOV. As illustrated in Fig.~\ref{pic_sim1}(c), our Follow Everything enables the robot to rapidly approach the leader’s last known pose at maximum speed, thereby minimizing the duration of disconnection and achieving the highest follow success rate of 96.9\%. In comparison, the ablated variant FE-N-GP does not employ a graph-based planner, nor does it consider the leader’s final orientation or use a speed-maximizing strategy. As a result, its follow success rate drops to 81.3\%. When it reaches the leader’s last known pose, the leader has moved far away, causing the leader to be forever lost. 
In methods such as Alaa and FE-N-DFB, which rely solely on the temporal memory buffer, as the leader often becomes partially visible in the final frames prior to disappearance, these incomplete embeddings may mislead the model and prevent accurate matching with the current scene, resulting in re-identification success rates of only 21.8\% and 62.5\%, respectively. In contrast, Follow Everything leverages the distance frame buffer to mitigate this issue, enabling robust leader re-identification.

\textbf{UAV leader in forest}. In this scenario, multiple randomly generated cylindrical obstacles are placed to simulate trees in a forest, and the leader is a small UAV. The main challenge lies in balancing following the leader and avoiding obstacles.

\begin{figure}[!t]
	\centering
	\includegraphics[width=3.4in]{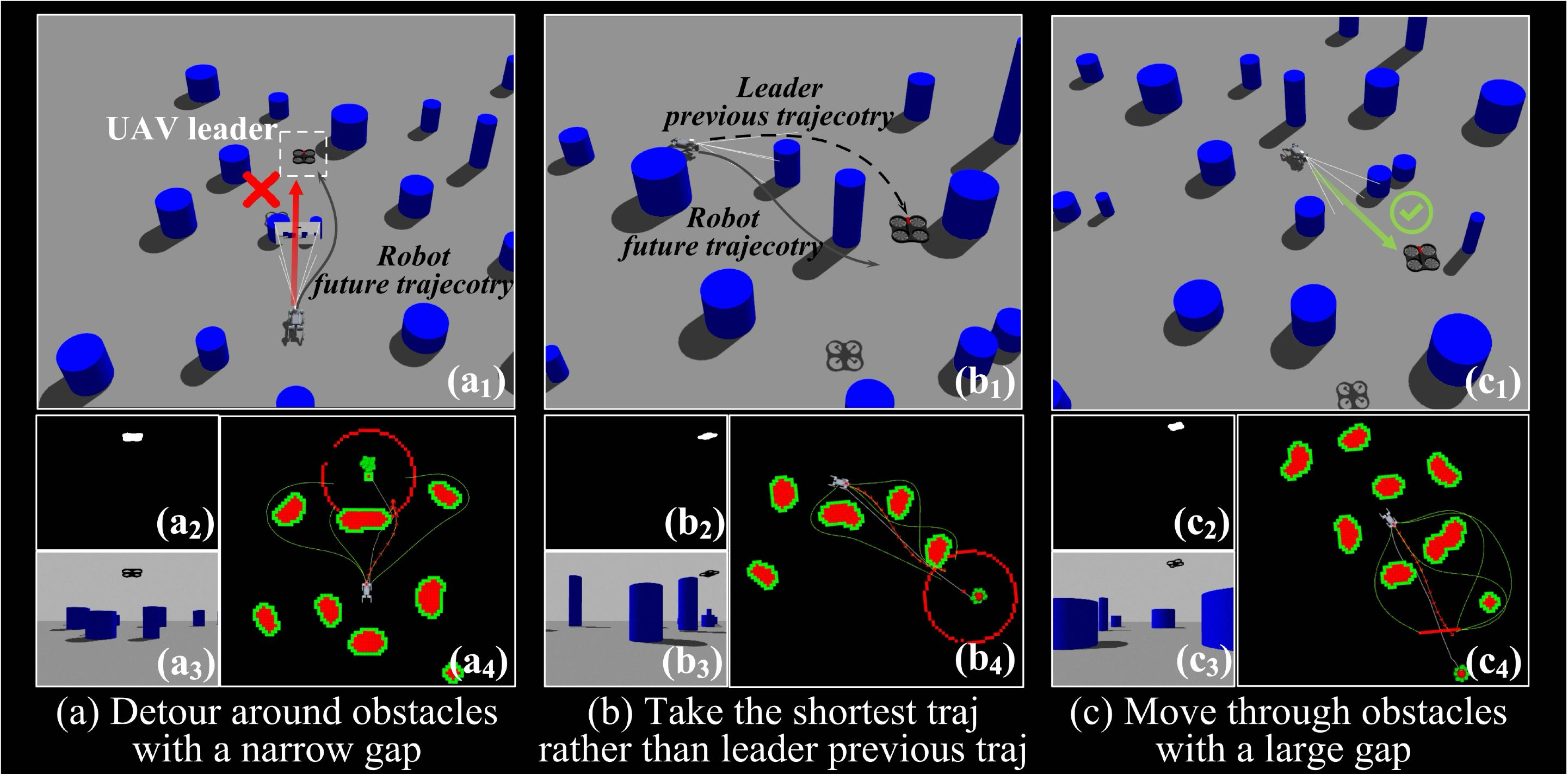}
	\caption{Simulation for following a UAV in a forest. (a) After the UAV flies through a narrow gap, the robot detours for safety. (b) Rather than replicating the UAV’s historical trajectory, the robot identifies a shorter one. (c) When the gap is wide enough, the robot passes through it for efficiency.
    }
	\label{pic_sim2}
\end{figure}

When the UAV flies over two closely spaced obstacles, as shown in Fig.~\ref{pic_sim2}(a), the gap between them is too narrow for the follower to pass through. So our Follow Everything enables the robot to promptly detour around the obstacles. During this process, the UAV may temporarily disappear from the robot’s FOV. While the goal-state adaptation transitions from the following state to the planning state, the graph-based planner ensures behavioral consistency by switching the robot’s goal from a circle set to a point. 

\begin{table*}[t]
\centering
\caption{Comparison of our Follow Everything to baselines and ablations. Our method achieves the best performance in all metrics}
\begin{tabular}{lcccc}
\hline
                       & Follow success rate [$\uparrow$] \hspace{6pt} & Average leader loss time ratio [$\downarrow$] \hspace{6pt} & Collision rate [$\downarrow$] \hspace{6pt} & Average distance [$\downarrow$] \hspace{6pt} \\ \hline
Alaa (baseline)~\cite{2024_follow_anything}                   & 21.8\%              & 23.8\%                         & 66.9\%         & 3.3 m            \\ \hline
FE-N-DFB (ablation)    & 62.5\%              & 44.0\%                         & 25.0\%         & 3.1 m            \\ \hline
FE-N-GP (ablation)     & 81.3\%              & 20.0\%                         & 16.3\%         & 2.6 m            \\ \hline
Follow Everything (ours) \hspace{6pt} & \textbf{96.9\%}              & \textbf{10.7\%}                         & \textbf{1.8\%}          & \textbf{2.0 m}          \\ \hline 
\label{table}
\end{tabular}
\end{table*}

\begin{figure}[!t]
	\centering
	\includegraphics[width=3.2in]{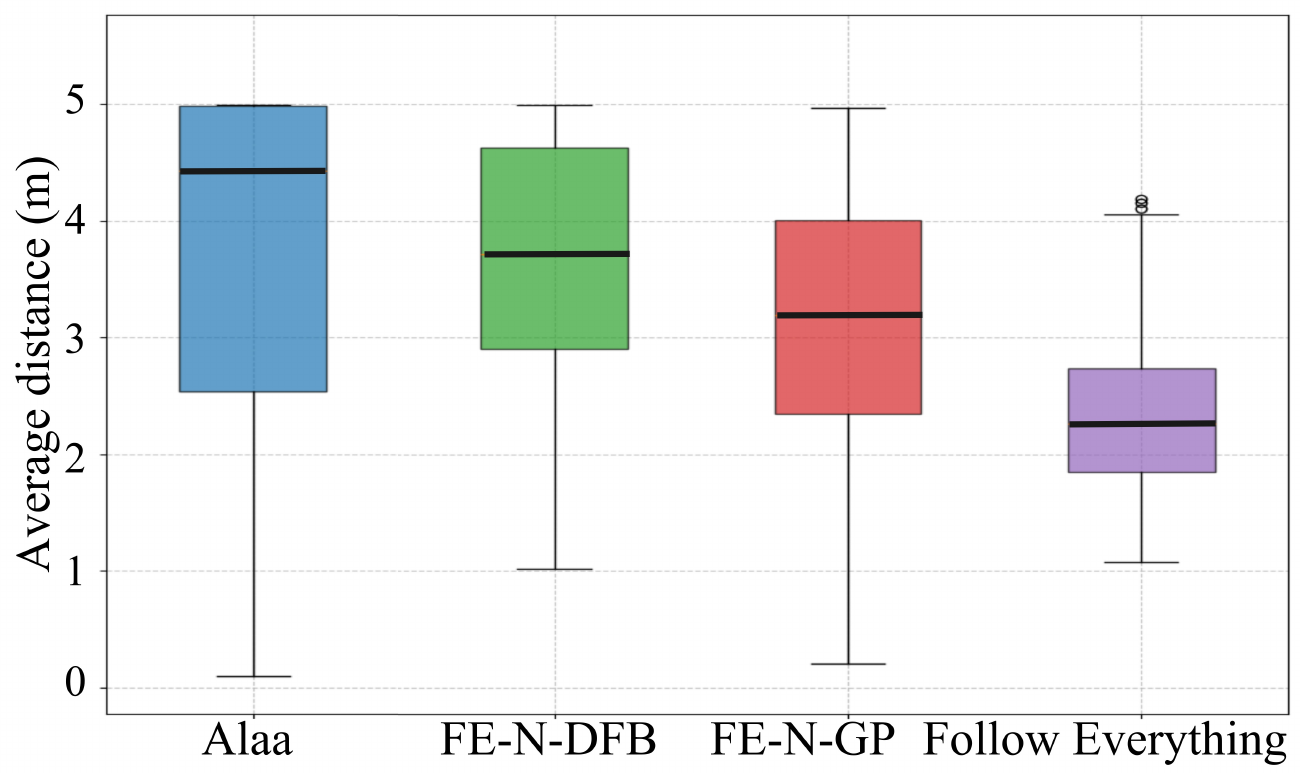}
	\caption{Average robot-leader distance in 160 tests: Alaa 3.3 m, FE-N-DFB 3.1m, FE-N-GP 2.6 m, and Follow Everything 2.0 m.}
	\label{box_sim3}
\end{figure}

After the UAV detours around two obstacles, as shown in Fig.~\ref{pic_sim2}(b), the graph-based planner generates multiple feasible trajectories and selects the fastest one, rather than replicating the UAV’s historical path. This strategy helps the robot maintain a close distance to the leader. As shown in Table.~\ref{table} and Fig.~\ref{box_sim3}, our method achieves the lowest average distance of 2.0 meters and robustly maintains it with the smallest variance, which reduces the likelihood of the leader leaving the robot’s FOV (too close can cause the UAV to leave the robot’s view, too far makes it hard to recognize), thus improving the overall follow success rate.

When the gap between obstacles is sufficiently wide for the robot to pass through, as shown in Fig.~\ref{pic_sim2}(c), the robot directly moves through it in order to maximize efficiency.

\textbf{Pedestrian leader in factory}. In this scenario, the leader is a pedestrian, and the environment contains various obstacles with diverse shapes, sizes, and heights. As shown in Fig.~\ref{pic_sim3}(a), due to the relatively large size of the leader, it is frequently partially observed by the robot, as shown in (b). This partial observation, combined with frequent visibility loss during obstacle avoidance and the visual similarity between parts of the obstacles and parts of the pedestrian’s body, pose a great challenge for the leader detection/segmentation.

Fig.~\ref{pic_sim3}(c) demonstrates the performance of the switching state. Upon receiving the prompt "follow the pedestrian on the left", the robot switches its target to a different pedestrian far away from it. Given the large distance between the robot and the new leader, the chasing state is activated. The robot moves at its maximum speed and passes underneath a workstation to rapidly approach the new leader.

Fig.~\ref{pic_sim3}(d) illustrates the retreating state, where the leader actively steps backward. The robot retreats accordingly while avoiding nearby construction cones, ensuring safety. As a result, Follow Everything achieves a collision rate of only 1.8\%, which is substantially lower than that of other methods. A key perception challenge here is that when the leader is close to the robot, only a portion of the leader's body is visible. When the leader starts moving forward again, SAM2 without a distance frame buffer often fails to re-identify the leader, leading to a collision rate of 66.9\% in Alaa. and 25.0\% in FE-N-DFB.
Fig.~\ref{pic_sim3}(e) shows how the planning state contributes to maintaining robust following behavior after the leader turns a corner and temporarily disappears from the robot’s FOV.

\begin{figure}[!t]
	\centering
	\includegraphics[width=3.4in]{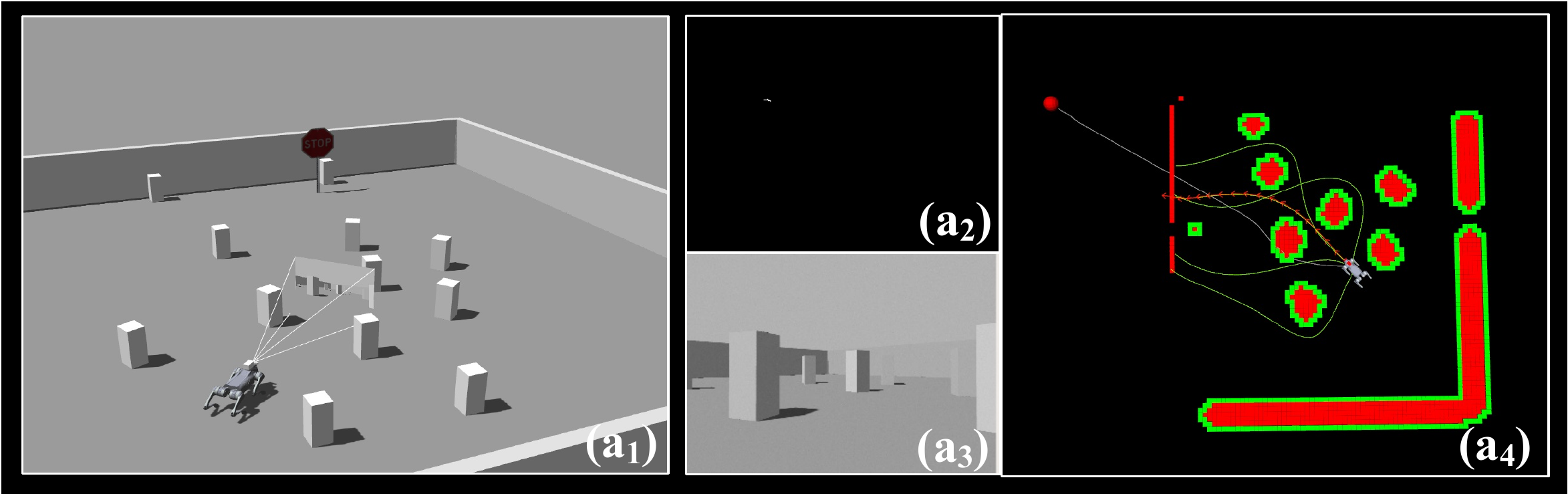}
	\caption{Simulation for following a stop sign. The graph-based planner provides multiple candidate trajectories so that a safer one can be identified. }
	\label{pic_sim4}
\end{figure}

\textbf{Stop sign leader in dynamic scenario}. As shown in Fig. \ref{pic_sim4}, 
in this scenario, the leader is a stop sign, and dynamic obstacles move randomly, changing their velocities every 3 seconds. The graph-based planner enables the generation of multiple candidate trajectories, ensuring that a safer trajectory is not only available but also selected. This capability allows Follow Everything to significantly outperform other methods, achieving the lowest collision rate of 1.8\%.

\section{Real-world Experiment}
In the experiment, the same legged robot is used as the follower, with a person or a wheel-legged robot as the leader. The experiment is conducted in both indoor and outdoor environments. A laptop with Intel i7 and Nvidia RTX 3070 is mounted on the legged robot as the onboard computer.

\begin{figure*}[!t]
	\centering
	\includegraphics[width=6.9in]{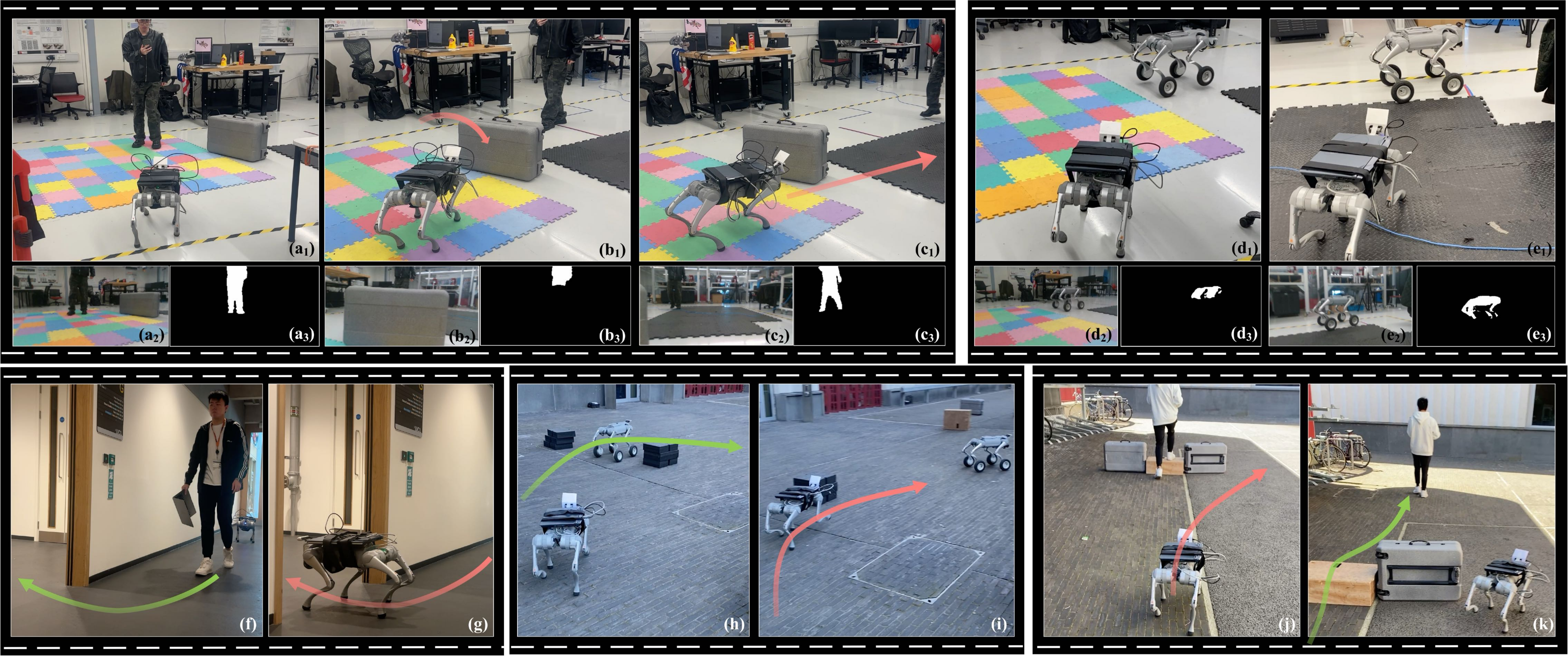}
	\caption{Illustration of real-world experiments. (a-c) While following a pedestrian, an appropriate safe distance allows the robot to rotate in place and avoid unnecessary detours. (d-e) The robot leader is reliably identified and followed. (f-g) The planning state allows the robot to follow and re-identify the leader after moving out of the room. (h-i) The robot takes a shorter path to quickly approach the leader. (j-k) The alternation between following and planning states enables the robot to follow the leader while avoiding obstacles when the leader steps over a box. }
	\label{pic_exp}
\end{figure*}

\textbf{Following state in indoor scenario}. As shown in Fig.~\ref{pic_exp}(a), a suitcase is placed in the scene center, and the robot first follows the pedestrian, maintaining a relatively stable safe distance. As shown in Fig.~\ref{pic_exp}(b), during the pedestrian’s clockwise detour around the obstacle, the robot continuously stays in the following state, adjusting its orientation to always face the leader while maintaining a safe distance, thus avoiding unnecessary detours. As shown in Fig.~\ref{pic_exp}(c), once the pedestrian moves towards a further location, the robot detours counterclockwise around the obstacle and continues to follow the pedestrian. As shown in Fig.~\ref{pic_exp}(d-e), when the pedestrian leader is replaced by a legged robot with wheels, the robot’s following behavior remains stable.

\textbf{Planning state in indoor scenario}. As shown in Fig.~\ref{pic_exp}(f-g), the robot initially follows the pedestrian’s movement. Once the person exits the room, the planning state is activated, and the robot moves to the last known position of the leader and successfully re-identifies the leader.

\textbf{Following state in outdoor scenario}. As shown in Fig.~\ref{pic_exp}(h-i), a wheeled legged robot leader moves through obstacles and then turns right. By maintaining a safe distance from the leader, the robot avoids replicating the leader’s path and instead turns right to quickly approach the leader.

\textbf{Planning state in outdoor scenario}. As shown in Fig.~\ref{pic_exp}(j-k), the leader is a pedestrian, moving in front of the robot and then stepping over a box. To avoid collision, the following state enables the robot to turn right and detour around the obstacles. During this process, the pedestrian temporarily leaves the robot’s FOV, and the planning state ensures the robot continues detouring until the leader reappears in the robot’s FOV, triggering the following state again.

\section{Conclusion}
This paper proposes a unified framework for robust leader-following, which integrates a segmentation model with a distance frame buffer to enable tracking of arbitrary leaders, along with a goal-aware adaptation module and a graph-based planner for dynamic parameter adjustment, goal state decision-making and trajectory generation. Simulations and indoor/outdoor experiments with a legged robot follower and diverse leaders verify that the framework significantly boosts following success rates, shortens leader visual loss duration, lowers collision rates, and reduces the average follower-leader distance.

\bibliographystyle{IEEEtran}
\bibliography{./IEEEfull}

@article{wang2025trackvla,
  title={Trackvla: Embodied visual tracking in the wild},
  author={Wang, Shaoan and Zhang, Jiazhao and Li, Minghan and Liu, Jiahang and Li, Anqi and Wu, Kui and Zhong, Fangwei and Yu, Junzhi and Zhang, Zhizheng and Wang, He},
  journal={arXiv preprint arXiv:2505.23189},
  year={2025}
}

@inproceedings{rpl1,
  title={Dippest: Diffusion-based path planner for synergistic trajectory generation applied on quadrupedal robots},
  author={Stamatopoulou, M and Liu, J and Kanoulas, D},
  booktitle={IEEE/RSJ IROS},
  year={2024}
}

@inproceedings{rpl2,
  title={Dipper: Diffusion-based 2d path planner applied on legged robots},
  author={Liu, Jianwei and Stamatopoulou, Maria and Kanoulas, Dimitrios},
  booktitle={2024 IEEE International Conference on Robotics and Automation (ICRA)},
  pages={9264--9270},
  year={2024},
  organization={IEEE}
}

@misc{rpl3,
      title={Local Path Planning among Pushable Objects based on Reinforcement Learning}, 
      author={Linghong Yao and Valerio Modugno and Andromachi Maria Delfaki and Yuanchang Liu and Danail Stoyanov and Dimitrios Kanoulas},
      year={2024},
      eprint={2303.02407},
      archivePrefix={arXiv},
      primaryClass={cs.RO},
}

@misc{rpl4,
      title={SanD-Planner: Sample-Efficient Diffusion Planner in B-Spline Space for Robust Local Navigation}, 
      author={Jincheng Wang and Lingfan Bao and Tong Yang and Diego Martinez Plasencia and Jianhao Jiao and Dimitrios Kanoulas},
      year={2026},
      eprint={2602.00923},
      archivePrefix={arXiv},
      primaryClass={cs.RO},
}

@ARTICLE{10909198,
  author={Roy, Liam and Croft, Elizabeth A. and Ramirez, Alex and Kulić, Dana},
  journal={IEEE Robotics and Automation Letters}, 
  title={GPT-Driven Gestures: Leveraging Large Language Models to Generate Expressive Robot Motion for Enhanced Human-Robot Interaction}, 
  year={2025},
  volume={10},
  number={5},
  pages={4172-4179},
  doi={10.1109/LRA.2025.3547631}
}

@article{2023_yolo,
  title={A comprehensive review of yolo architectures in computer vision: From yolov1 to yolov8 and yolo-nas},
  author={Terven, Juan and C{\'o}rdova-Esparza, Diana-Margarita and Romero-Gonz{\'a}lez, Julio-Alejandro},
  journal={Machine learning and knowledge extraction},
  volume={5},
  number={4},
  pages={1680--1716},
  year={2023},
  publisher={MDPI}
}

@inproceedings{2021_lidar,
  title={Center-based 3d object detection and tracking},
  author={Yin, Tianwei and Zhou, Xingyi and Krahenbuhl, Philipp},
  booktitle={Proceedings of the IEEE/CVF conference on computer vision and pattern recognition},
  pages={11784--11793},
  year={2021}
}

@article{2023_multimodal,
  title={Hybrid object detection and tracking for cooperative perception using 3D LiDAR},
  author={Meng, Zonglin and Xia, Xin and Xu, Runsheng and Liu, Wei and Ma, Jiaqi},
  journal={IEEE Transactions on Intelligent Vehicles},
  volume={8},
  number={8},
  pages={4069--4080},
  year={2023},
  publisher={IEEE}
}

@INPROCEEDINGS{2015_kalman_filter,
  author={Li, Qiang and Li, Ranyang and Ji, Kaifan and Dai, Wei},
  booktitle={2015 8th International Conference on Intelligent Networks and Intelligent Systems (ICINIS)}, 
  title={Kalman Filter and Its Application}, 
  year={2015},
  volume={},
  number={},
  pages={74-77},
  doi={10.1109/ICINIS.2015.35}
}

@article{2024_follow_anything,
  title={Follow anything: Open-set detection, tracking, and following in real-time},
  author={Maalouf, Alaa and Jadhav, Ninad and Jatavallabhula, Krishna Murthy and Chahine, Makram and Vogt, Daniel M and Wood, Robert J and Torralba, Antonio and Rus, Daniela},
  journal={IEEE Robotics and Automation Letters},
  volume={9},
  number={4},
  pages={3283--3290},
  year={2024},
  publisher={IEEE}
}

@inproceedings{2021_ultrasonic,
  title={Human following robot using ultrasonic sensor},
  author={Tripathi, Abhinav and Khan, Mohd Ammar and Pandey, Akash and Yadav, Pankaj and Sharma, Amit Kumar},
  booktitle={2021 3rd International Conference on Advances in Computing, Communication Control and Networking (ICAC3N)},
  pages={764--770},
  year={2021},
  organization={IEEE}
}

@article{2022_laser,
  title={A quadruped robot obstacle avoidance and personnel following strategy based on ultra-wideband and three-dimensional laser radar},
  author={Li, Zhi and Li, Bin and Liang, Qixing and Liu, Weilong and Hou, Landong and Rong, Xuewen},
  journal={International Journal of Advanced Robotic Systems},
  volume={19},
  number={4},
  pages={17298806221114705},
  year={2022},
  publisher={SAGE Publications Sage UK: London, England}
}

@article{2023_wave,
  title={Efficient volumetric mapping of multi-scale environments using wavelet-based compression},
  author={Reijgwart, Victor and Cadena, Cesar and Siegwart, Roland and Ott, Lionel},
  journal={arXiv preprint arXiv:2306.01279},
  year={2023}
}

@article{2024_wave,
  title={Obstacle-avoidant leader following with a quadruped robot},
  author={Scheidemann, Carmen and Werner, Lennart and Reijgwart, Victor and Cramariuc, Andrei and Chomarat, Joris and Chiu, Jia-Ruei and Hutter, Marco},
  journal={arXiv preprint arXiv:2410.00572},
  year={2024}
}

@ARTICLE{2025_MPC_TRO,
  author={Benders, Dennis and Köhler, Johannes and Niesten, Thijs and Babuška, Robert and Alonso-Mora, Javier and Ferranti, Laura},
  journal={IEEE Transactions on Robotics}, 
  title={Embedded Hierarchical MPC for Autonomous Navigation}, 
  year={2025},
  volume={41},
  number={},
  pages={3556-3574},
  doi={10.1109/TRO.2025.3567529}
}

@INPROCEEDINGS{gaze,
  author={Zhang, Qianyi and Hu, Zhengxi and Song, Yinuo and Pei, Jiayi and Liu, Jingtai},
  booktitle={2023 IEEE International Conference on Robotics and Automation (ICRA)}, 
  title={The Human Gaze Helps Robots Run Bravely and Efficiently in Crowds}, 
  year={2023},
  volume={},
  number={},
  pages={7540-7546},
  doi={10.1109/ICRA48891.2023.10161222}
}

@article{han2025neupan,
  title={Direct point robot navigation with end-to-end model-based learning},
  author={Han, Ruihua and Wang, Shuai and Wang, Shuaijun and Zhang, Zeqing and Chen, Jianjun and Lin, Shijie and Li, Chengyang and Xu, Chengzhong and Eldar, Yonina C and Hao, Qi and others},
  journal={IEEE Transactions on Robotics},
  year={2025},
  publisher={IEEE}
}

@article{zhang2021efficient,
  title={Efficient motion planning based on kinodynamic model for quadruped robots following persons in confined spaces},
  author={Zhang, Zhen and Yan, Jiaqing and Kong, Xin and Zhai, Guangyao and Liu, Yong},
  journal={IEEE/ASME Transactions on Mechatronics},
  volume={26},
  number={4},
  pages={1997--2006},
  year={2021},
  publisher={IEEE}
}

@article{2024_STCTEB,
  title={STC-TEB: Spatial-Temporally Complete Trajectory Generation Based on Incremental Optimization},
  author={Zhu, Zeqing and Zhang, Qianyi and Song, Yinuo and Yang, Yifan and Liu, Jingtai},
  journal={IEEE Robotics and Automation Letters},
  year={2024},
  publisher={IEEE}
}

@misc{2025_GATEB,
      title={GA-TEB: Goal-Adaptive Framework for Efficient Navigation Based on Goal Lines}, 
      author={Qianyi Zhang and Wentao Luo and Ziyang Zhang and Yaoyuan Wang and Jingtai Liu},
      year={2024},
      eprint={2409.10009},
      archivePrefix={arXiv},
      primaryClass={cs.RO},
}

@article{2024_evf_sam2,
  title={Early vision-language fusion for text-prompted segment anything model},
  author={Zhang, Yuxuan and Cheng, Tianheng and Hu, Rui and Liu, Lei and Liu, Heng and Ran, Longjin and Chen, Xiaoxin and Liu, Wenyu and Wang, Xinggang},
  journal={arXiv preprint arXiv:2406.20076},
  year={2024}
}

@ARTICLE{2025_tro_homo,
  author={de Groot, Oscar and Ferranti, Laura and Gavrila, Dariu M. and Alonso-Mora, Javier},
  journal={IEEE Transactions on Robotics}, 
  title={Topology-Driven Parallel Trajectory Optimization in Dynamic Environments}, 
  year={2025},
  volume={41},
  number={},
  pages={110-126},
  doi={10.1109/TRO.2024.3475047}
}

@inproceedings{2010_homo,
  title={Search-based path planning with homotopy class constraints},
  author={Bhattacharya, Subhrajit},
  booktitle={Proceedings of the AAAI conference on artificial intelligence},
  volume={24},
  number={1},
  pages={1230--1237},
  year={2010}
}

@article{2017_RAS_TEB,
title = {Integrated online trajectory planning and optimization in distinctive topologies},
journal = {Robotics and Autonomous Systems},
volume = {88},
pages = {142-153},
year = {2017},
issn = {0921-8890},
doi = {https://doi.org/10.1016/j.robot.2016.11.007},
author = {Christoph Rösmann and Frank Hoffmann and Torsten Bertram},
}

@INPROCEEDINGS{LiuDiPPeR2024,
  author={Liu, Jianwei and Stamatopoulou, Maria and Kanoulas, Dimitrios},
  booktitle={IEEE International Conference on Robotics and Automation (ICRA)}, 
  title={Diffusion-based 2D Path Planner applied on Legged Robots}, 
  year={2024},
  pages={9264-9270},
  doi={10.1109/ICRA57147.2024.10610013}}

\vfil

\end{document}